\documentclass[journal]{IEEEtran}
\usepackage[switch]{lineno}
\let\linenumbers\nolinenumbers\nolinenumbers

\usepackage{cite}
\usepackage[utf8]{inputenc} %
\usepackage[T1]{fontenc}    %
\usepackage{hyperref}
\usepackage{url}
\usepackage{booktabs}       %
\usepackage{amsfonts}       %
\usepackage{nicefrac}       %
\usepackage{microtype}      %
\usepackage{xcolor}
\usepackage{graphicx}
\usepackage{wrapfig}
\usepackage{enumitem}
\usepackage{comment}
\usepackage{tipa}
\usepackage{amsmath}
\usepackage{listings}
\usepackage{selectp}
\usepackage{multirow}
\usepackage[frozencache]{minted}
\usepackage{bbm}
\usepackage{bm}
\usepackage{amssymb}
\usepackage{graphicx}
\usepackage{subfig}
\usepackage{mathptmx}
\usepackage{adjustbox}

\definecolor{blue-violet}{rgb}{0.54, 0.17, 0.89}

\ifCLASSINFOpdf
\else
\fi

\newcommand{\comb}[2]{{}_{#1}\mathrm{C}_{#2}}

\begin{document}
\linenumbers
%
\title{Contrastive Multiview Coding with Electro-optics

for SAR Semantic Segmentation}
%
%
%

\author{Keumgang~Cha,
        Junghoon~Seo,
        and~Yeji~Choi,~\IEEEmembership{Member,~IEEE}
\thanks{Keumgang Cha, Junghoon Seo, and Yeji Choi are with SI-Analytics, Daejeon 34051, South Korea (e-mail: chagmgang@si-analytics.ai; jhseo@si-analytics.ai; yejichoi@si-analytics.ai).}
}

%
%

\markboth{IEEE Geoscience and Remote Sensing Letters}%
{Shell \MakeLowercase{\textit{et al.}}: Bare Demo of IEEEtran.cls for Journals}
%



\maketitle

\begin{abstract}
In the training of deep learning models, how the model parameters are initialized greatly affects the model performance, sample efficiency, and convergence speed.
Representation learning for model initialization has recently been actively studied in the remote sensing field.
In particular, the appearance characteristics of the imagery obtained using the a synthetic aperture radar (SAR) sensor are quite different from those of general electro-optical (EO) images, and thus representation learning is even more important in remote sensing domain.
Motivated from contrastive multiview coding, we propose multi-modal representation learning for SAR semantic segmentation.
Unlike previous studies, our method jointly uses EO imagery, SAR imagery, and a label mask.
Several experiments show that our approach is superior to the existing methods in model performance, sample efficiency, and convergence speed.
\end{abstract}


%
\IEEEpeerreviewmaketitle

\section{Introduction}
	\IEEEPARstart{T}He application of deep learning for synthetic aperture radar (SAR) is receiving increased attention. As one reason for this, unlike electro-optical (EO) sensors, SAR sensors can capture images without being affected by flight altitude and weather. Many studies have attempted to solve the terrain surface classification, object detection, parameter inversion, and despeckling in the SAR domain with deep learning \cite{zhu2021deep}.
Considering this trend, it is worth overcoming a critical issue to effectively train deep neural networks for SAR imagery.

One of the critical factors in training a deep neural network is how to initialize the model parameters.
For remote sensing applications on EO images, transfer learning initialized from the other generic large-scale tasks is considered the dominant methodology.
For instance, transfer learning from the ImageNet-pretrained model has been utilized in several EO remote sensing tasks, such as land cover classification \cite{scott2017training}, disaster damage detection \cite{seo2019revisiting}, cloud detection \cite{9314019}, building segmentation \cite{9416747}, and object detection \cite{long2017accurate}.
This \textit{de facto} initialization is known to reduce the number of data required for training, enable faster training, and increase the stability of the training procedure.
However, transfer learning is also known to be less effective as the difference between the source and target domains increases \cite{neumann2020training,mensink2021factors}. This can be particularly problematic in the SAR domain, because SAR sensors have extremely different characteristics from those of general EO cameras.

In a similar vein, several recent studies \cite{ren2021mutual,jean2019tile2vec,kang2020deep,manas2021seasonal,10.1145/3357384.3358001,uzkent2019learning,ayush2020geography} have dealt with various types of representation learning in the remote sensing domain.
According to the number of modals used for representation learning, they can be divided into single-modal \cite{ren2021mutual,jean2019tile2vec,kang2020deep,manas2021seasonal} and multi-modal \cite{10.1145/3357384.3358001,uzkent2019learning,ayush2020geography} methodologies.

Because single-modal methods assume that only single domain data can be accessed, self-supervised learning methodologies such as SimCLR \cite{chen2020simple} or MoCo \cite{he2020momentum} are often used. By contrast, multimodal methods conduct representation learning using the relationships between a plurality of different domains. Unlike the single-modal method, which uses only a single domain of data for representation learning, the multimodal approach has the advantage of being able to implicitly make the most of important information shared by multiple domains. For example, in \cite{10.1145/3357384.3358001}, representations of regions are learned jointly from a satellite image, point-of-interest, and human mobility. In \cite{uzkent2019learning}, the authors utilized geo-referenced Wikipedia articles with satellite imagery of the corresponding locations. In \cite{ayush2020geography}, spatially aligned EO images taken over time are leveraged along with geo-location information. To obtain good pre-trained neural networks using multiple domain pairs, these approaches have adopted various objective functions of representation learning. These include auto-encoding loss \cite{10.1145/3357384.3358001}, cross-entropy loss \cite{10.1145/3357384.3358001,uzkent2019learning,ayush2020geography}, cosine similarity loss \cite{uzkent2019learning}, and temporal contrastive loss in a single domain \cite{ayush2020geography}. These existing studies mainly focused on using the relationship between a single satellite sensor and other meta-data not obtained from satellite sources.

\begin{figure}
 \centering
 \subfloat[SAR]{\label{fig:sar1}\includegraphics[width=0.095\textwidth]{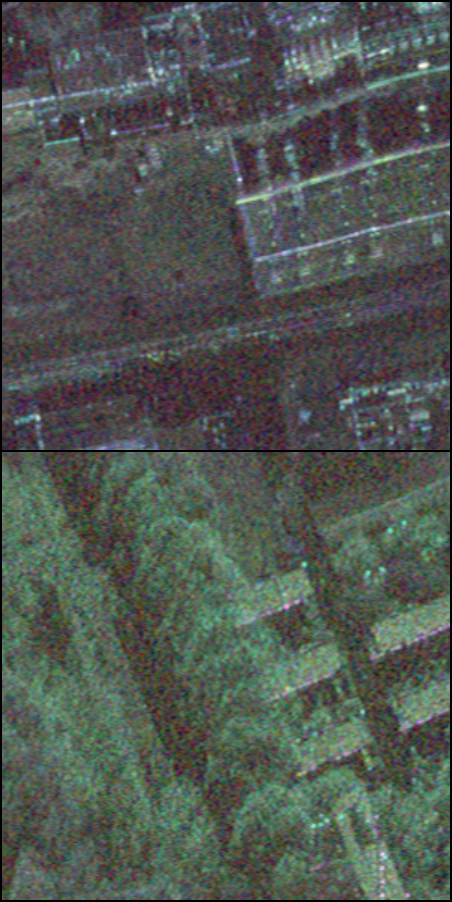}}\hfill
 \subfloat[GT]{\label{fig:gt1}\includegraphics[width=0.095\textwidth]{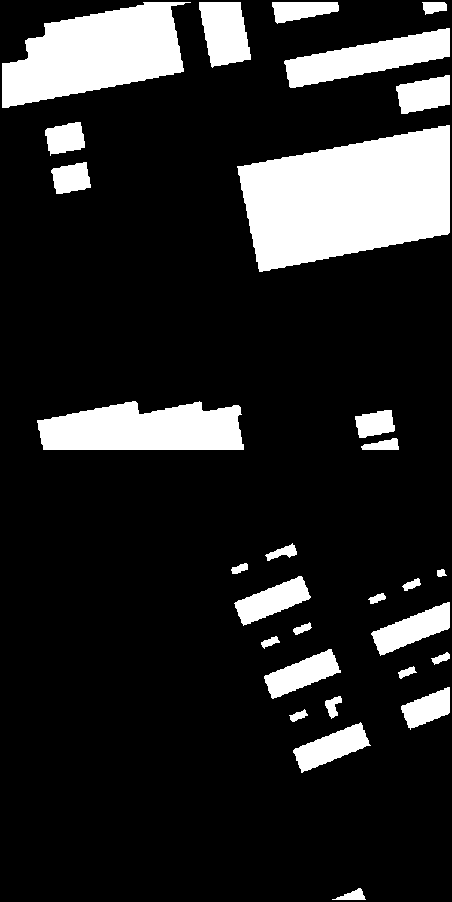}}\hfill
 \subfloat[Scratch]{\label{fig:scratch1}\includegraphics[width=0.095\textwidth]{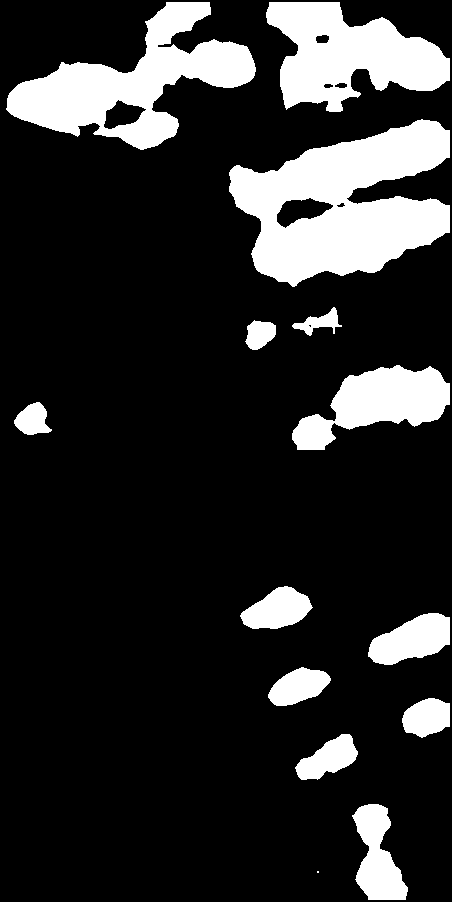}}\hfill
 \subfloat[ImageNet]{\label{fig:imagenet1}\includegraphics[width=0.095\textwidth]{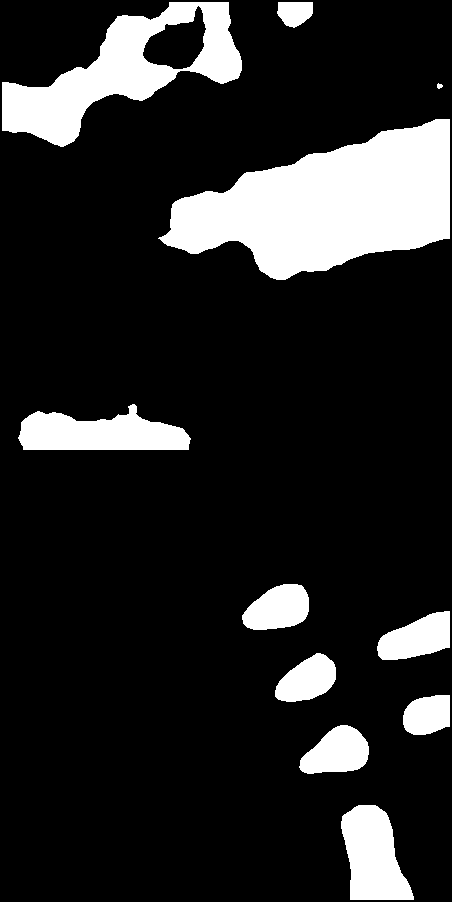}}\hfill
 \subfloat[Ours]{\label{fig:cmc1}\includegraphics[width=0.095\textwidth]{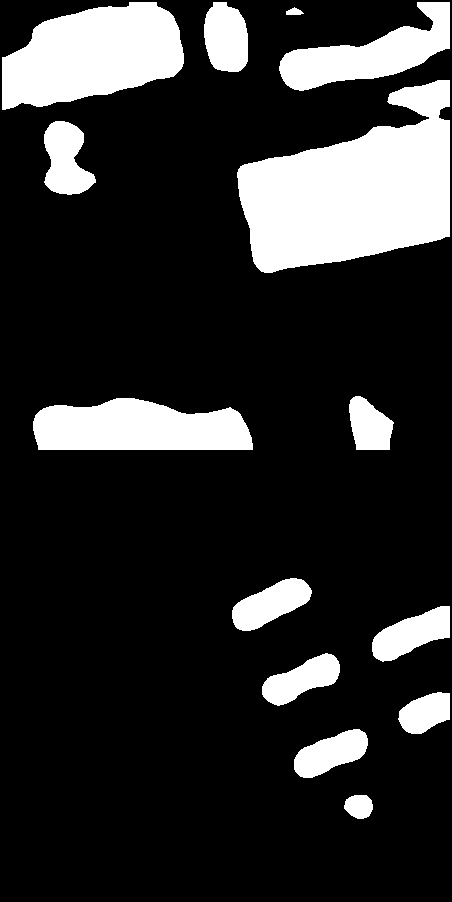}}
 \caption{Visual comparison on baselines and our approach. Inference results of DeepLabV3 building segmentation model by different weight initialization schemes for the same SAR image and same training steps are shown. The detailed experiment settings are described in Section \ref{sec:exp}.}
 \label{fig:animals}
\end{figure}

On the contrary, here our question is, \textit{“Can the auxiliary use of EO images help	data-efficient semantic segmentation in the SAR domain?”} This paper focuses on semantic segmentation in SAR images and presents a novel multi-modal representation learning based on the spatial consistency. In other words, our approach makes full use of the property in which image pairs of the same area with different modality sensors have spatially consistent features, whereas image pairs of different areas do not.
Our methodology is based on contrastive multiview coding \cite{tian2019contrastive}, which is a representation learning technique that models domain-invariant factors among multiple domains. Figure \ref{fig:animals} shows the effectiveness of the methodology covered in this study. Specifically, our contributions described in this paper are as follows:
\begin{itemize}
  \item We formulate multi-modal representation learning in contrastive multiview coding by considering the spatial nature of the three modalities. To the best of our knowledge, this is the first multi-modal approach in remote sensing field to use another satellite sensor for pretraining, which improves the performance of the SAR semantic segmentation model.
  \item We experimented on the possible combinations of these three modalities in detail. The empirical results indicate that our method is superior to other methodologies in terms of convergence speed, the final model performance, and data efficiency. Moreover, the effectiveness of our approach appears to be experimentally agnostic to the selection of the semantic segmentation model.
\end{itemize}

\section{Method}

\subsection{Contrastive Learning}

Contrastive learning \cite{henaff2020data} is learning method used to create a good representation by maximizing mutual information between differently augmented images from the same scene, i.e., \(\tilde{x}_i\), \(\tilde{x}_j\) which is called a positive pair. A convolution neural network, which is base encoder $f(*)$, extracts the representation of differently augmented images, that is, $h_i = f(\tilde{x}_i) $. A simple neural network, projection \(g(*)\), transforms representations into a space where contrastive loss is applied, i.e. $z_i = g(h_i) = g(f({x}_i))$.

A contrastive loss function is defined for a contrastive prediction. Let the similarity be cosine similarity, that is  $d(u,v)={u^T}v / ||u||||v||$. Then, the single loss function for one positive pair of $(i, j)$ is then written as,

\begin{equation}
\label{eq:cl}
l_{i,j} = -\log \frac{\exp(d(z_i,z_j) / \tau)}{\sum_{k=1}^{2N} \mathbbm{1}_{[k \neq i]} \exp(d(z_i,z_k) / \tau)},
\end{equation}
where $\mathbbm{1}_{[k \neq i]} $ is an indicator function evaluating to 1 iff $k \neq i$, $\tau$ is a temperature parameter and $N$ is a mini-batch size. The overall loss is the summation across all positive pairs, both $(i,j)$ and $(j,i)$, in mini-batch.

\begin{figure}[t]%
    \centering
    \includegraphics[width=0.42\textwidth]{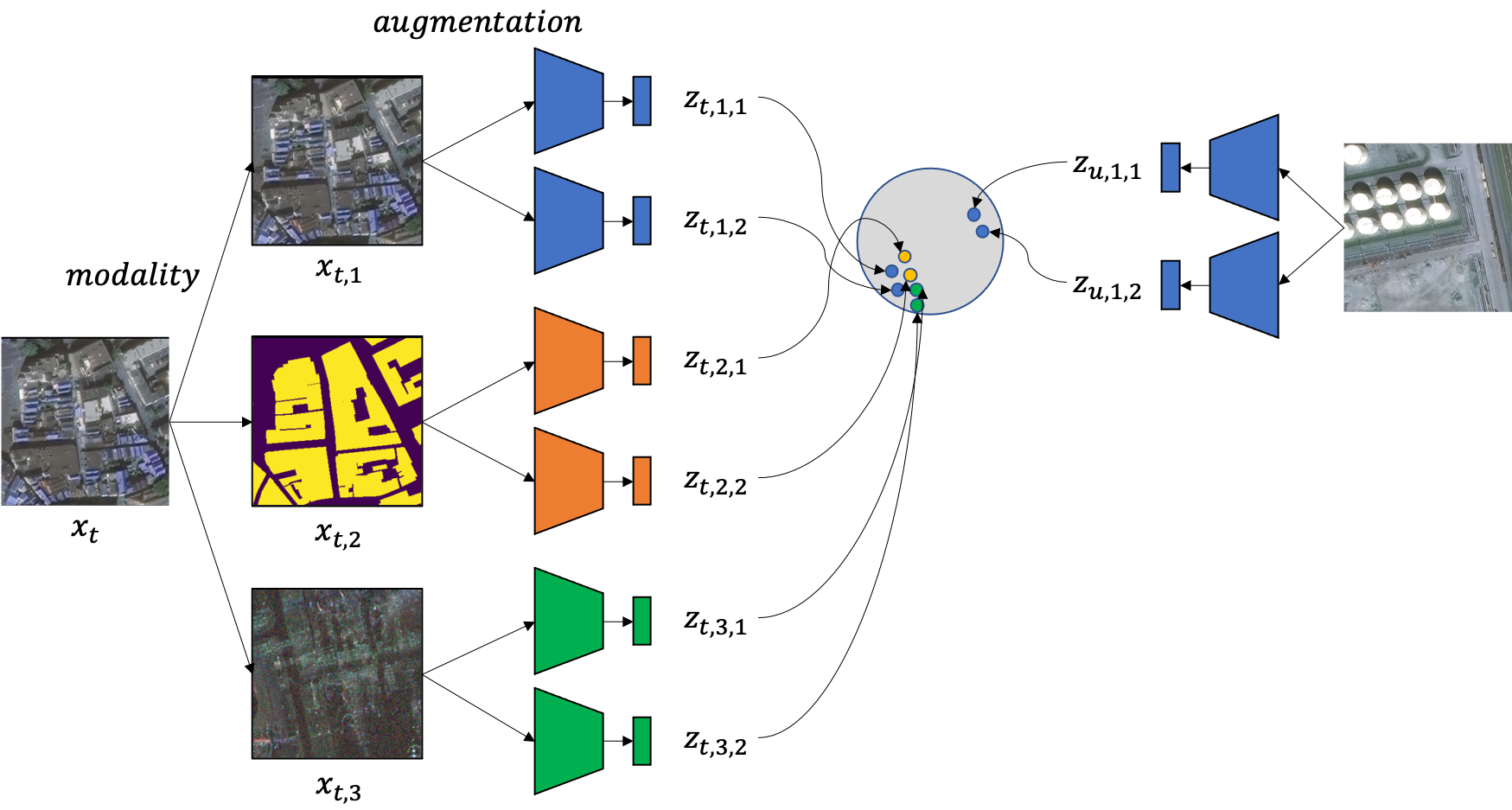}
    \caption{Contrastive multiview coding framework. Each scene, $x_t$, is sampled as a mini-batch, and modalities corresponding to the scene are selected. After augmentation, the latent, $z$, is generated through the base encoder and projection corresponding to each modality.}%
    \label{fig:cmc}%
\end{figure}

\subsection{Contrastive Multiview Coding}

Contrastive multiview coding \cite{tian2019contrastive} (CMC) is a multiview version of contrastive learning. In this letter, the multiview is constructed using different modalities and augmentations. If $M$ modalities and $N$ augmentations are applied to a single $t$-th scene, $x_t$, we can obtain $M \times N$ images in one scene, denoted as $x_{t,m,n}$. Although this method is quite similar to with contrastive learning, but the convolution neural network, which is base encoder $f(*)$, and simple neural network, which is projection $g(*)$, are configured to have the sample structure and different weights according to each modality, that is, $z_{t,i,j} = g_{\theta_i}(f_{\theta_i}(x_{t,i,j}))$. There are two optimization methods for contrastive multiview coding, core and full graph \cite{tian2019contrastive}. In this study, the full graph optimization is used to consider as much modality as possible. The full graph optimization is derived as below:
\begin{equation}
\label{eq:cmc}
    {L = \sum_{t,m,n} \mathbbm{1}_{[m \neq j \, or \, n \neq k]}
    -\log \frac{\exp(d(z_{t,m,n}, z_{i,j,k})/\tau)}{\sum_{i=1}^{B} \mathbbm{1}_{[t \neq i]}
    \exp(d(z_{t,m,n}, z_{i,j,k})/\tau)}},
\end{equation}
where $t=1,...,B$, $m=1,...,M$, $n=1,...,N$, and, $B$ is the mini-batch size.
Compared to Equation \ref{eq:cl}, Equation \ref{eq:cmc} considers all pairs among all combinations consisting of $M$ modalities and $N$ augmentations, and $\comb{M \times N}{2}$ relationships are built for each $t$-th batch instance.
Using full graph optimization, the mutual information between different modalities can be maximized in same scene.
Naturally, in the stage of representation learning, contrastive multiview coding requires $M \times N$ times as much computation and training time as compared to contrastive learning using a single pair \cite{kang2020deep,manas2021seasonal}. Despite the additional computational burden, we will show that contrastive multiview coding has significant advantages in terms of the convergence speed and final model performance after representation learning is completed.

\subsection{Contrastive Multiview Coding in Remote Sensing}

It is simple to define a positive sample condition to obtain representation features by contrastive learning in a large-scale general image. Different view created in the sample scene becomes a positive sample condition. For instance, in contrastive learning, a specific image is defined as a scene, and a different view is generated by operating different augmentations in a single scene. In contrastive multiview coding, views from the same scenes are defined as positive samples such as contrastive learning. However, the method for generating various views is different from contrastive learning. Various views are created by a different sensors such as depth map, segmentation map, grayscale map, and optical flow map.

However, in remote sensing, it is challenging to apply contrastive multiview coding using a method applied in the field of computer vision. Although it can be obtained through augmentation, it is not easy to obtain a view with various modalities that satisfy the same time and place. Although this problem has not been completely solved, it is treated as a positive sample if the scene is captured in the same place. Because buildings are not moving objects like ships or cars, so they have relatively invariant properties over time. In addition, the images taking the same space with different sensors are also considered to be positive samples. For example, EO and SAR images that capture the same space are positive samples and a semantic segmentation map for the same space was also used as a positive sample.

\section{Experiment}
\label{sec:exp}
\subsection{Data Preparation}

We used the SpaceNet 6 dataset \cite{shermeyer2020spacenet} for building semantic segmentation. The SpaceNet 6 dataset is publicly available and provided by the SpaceNet; a nonprofit organization launched as an open innovation project for free satellite images with co-registered map features. This dataset is prepared to capture building footprints under all weather conditions for computer vision and deep learning applications by combining SAR and electro-optical images. The SAR image is from a quad-polarized X-band sensor with four polarizations (i.e., HH, HV, VH, and VV) provided by Capella Space, and the EO image from the Maxar WorldView 2 satellite. Each scene had a pixel resolution of $900 \times 900$ with a GSD of approximately 0.5m. The area of interest is centered over the largest port of Rotterdam, the Netherlands in Europe. The SAR, EO, and building footprint maps are contained in the training dataset, whereas the SAR image is contained only in the test dataset. Therefore, we randomly divided the training dataset at a ratio of 8:2 for the training and validation dataset.
To use the rich information in the SAR image, we stacked the SAR image with three polarizations (HH, VV, and HV) for generating three-channel images and used it as an input.

\subsection{Experimental Setup}

To find the good modality combination for obtaining good representation extractor, the ablation study was conducted under four combinations.
If only SAR images were used for pretraining, six positive samples from data augmentation were used. In case of \textit{EO + SAR}, EO and SAR images were used, with six positive samples and three augmentations applied, respectively. Also, in \textit{SAR + GT}, SAR images and segmentation maps were used. Finally, the setting of \textit{SAR + GT + EO} used two EO images, two SAR images, and two segmentation maps as one positive sample set.

\subsubsection{Backbone Architecture and Semantic Segmentation Models}
As shown in Figure \ref{fig:cmc}, the convolution neural network, a base encoder $f(*)$, was one of the most widely used residual-based architectures: ResNet50 \cite{he2016deep}, ResNext50 \cite{xie2017aggregated} or Resnest50 \cite{zhang2020resnest}. In addition, the simple neural network, projection $g(*)$, consisted of a linear layer with an output size of 2,048, batch normalization \cite{ioffe2015batch}, rectified linear units \cite{nair2010rectified}, and a final linear layer with output dimension 2,048. To confirm that consistent results were produced not only for the single model, but also for other segmentation architectures, the same experiments were conducted on three popular segmentation models: U-Net\cite{ronneberger2015u}, BiSeNet\cite{yu2018bisenet} and DeepLabV3\cite{chen2017rethinking}.

\subsubsection{Training Details}
\label{sec:training-details}

Because the original scene is too large for an the input image, the scene was patched by $300 \times 300$ with 150 strides. After patching the scene, each patch was resized to $448 \times 448$ as an input of the base encoder. The evaluation was conducted after inference in the patch and merging the building footprint into a single scene again.

Data augmentation for pretraining was as follows in order: resize to a ratio of 1.2 ($537 \times 537$), randomly crop to a fixed size ($448 \times 448$), horizontal flip with a probability of 0.5, vertical flip with a probability of 0.5, rotation from $-45$ to $45$, Gaussian blur with kernel size of 23, and apply a sigma range from $0.1$ to $0.2$ with a probability of 0.5. For pretraining, the batch size, $B$, was 126 for 500 epochs, and the temperature parameter($\tau$) was 0.1. For optimization, base learning rate was 0.1, and the weight decay coefficient was 0.0005. We used a stochastic gradient descent optimizer with a warm-up cosine decay scheduler \cite{loshchilov2016sgdr} by 10 warm-up epochs.

For building semantic segmentation, the data augmentation was sequentially defined as follows: resize to $448 \times 448$, and apply a horizontal flip with probability of 0.5, and a vertical flip with 0.5 probability. The segmentation models were trained with a batch size of 72 for 25 epochs. Furthermore, segmentation models were trained with stochastic gradient descent optimizer with warm-up cosine decay scheduler by 1 warm-up epochs. The base learning rate was 0.0075, and the weight decay was 0.0005.

\begin{table}[!b]
    \centering
    \begin{adjustbox}{width=0.36\textwidth}
    \begin{tabular}{l|l|c|c}
    \toprule
    Seg Architecture & Pretrain Source & Acc & Building IoU \\
    \midrule
    U-Net & (1) Random & $0.9317$ & $0$ \\
         &  (2) ImageNet & $0.9594$ & $0.5126$ \\
         &  (3) SAR & $0.9678$ & $0.5996$ \\
         &  (4) SAR+EO & $0.9724$ & $0.6517$ \\
         &  (5) SAR+GT & $0.9742$ & $0.6708$ \\
         &  (6) SAR+GT+EO & $\bm{0.9745}$ & $\bm{0.6738}$ \\
    \midrule
    BiSeNet & (1) Random & $0.9356$ & $0.1366$ \\
         &  (2) ImageNet & $0.9606$ & $0.5152$ \\
         &  (3) SAR & $0.9683$ & $0.6062$ \\
         &  (4) SAR+EO & $0.9733$ & $0.6607$ \\
         &  (5) SAR+GT & $0.9729$ & $0.6562$ \\
         &  (6) SAR+GT+EO & $\bm{0.9738}$ & $\bm{0.6647}$ \\
    \midrule
    DeepLabV3 & (1) Random & $0.9302$ & $0.2732$ \\
         & (2) ImageNet & $0.9502$ & $0.4896$ \\
         & (3) SAR & $0.9673$ & $0.6280$ \\
         & (4) SAR+EO & $0.9736$ & $0.6874$ \\
         & (5) SAR+GT & $0.9738$ & $0.6878$ \\
         & (6) SAR+GT+EO & $\bm{0.9738}$ & $\bm{0.6890}$ \\
    \midrule
    \end{tabular}
    \end{adjustbox}
    \caption{This table shows the performance of building segmentation according to the conditions of contrastive multiview coding by segmentation architecture. The same was applied to U-Net, BiSeNet and DeepLabV3 to confirm that contrastive multiview coding shows consistent results for various segmentation. Generally, the more modals of images used for contrastive multiview coding, the better the performance.}
    \label{tab:Result Table}
\end{table}

\subsection{Results and Discussion}
\begin{figure}[t]%
 \centering
 \subfloat{{\includegraphics[width=0.25\textwidth]{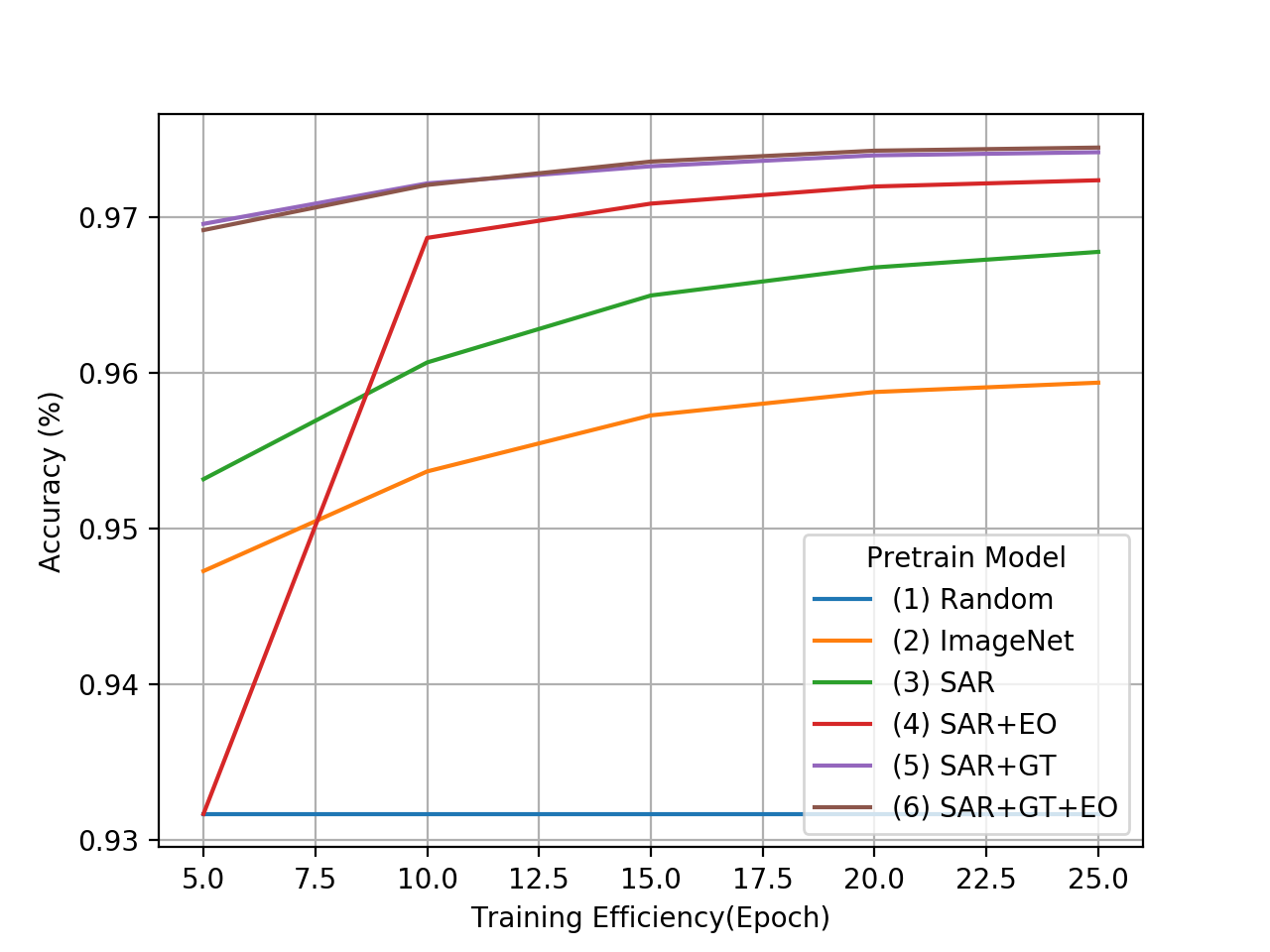} }}%
 \subfloat{{\includegraphics[width=0.25\textwidth]{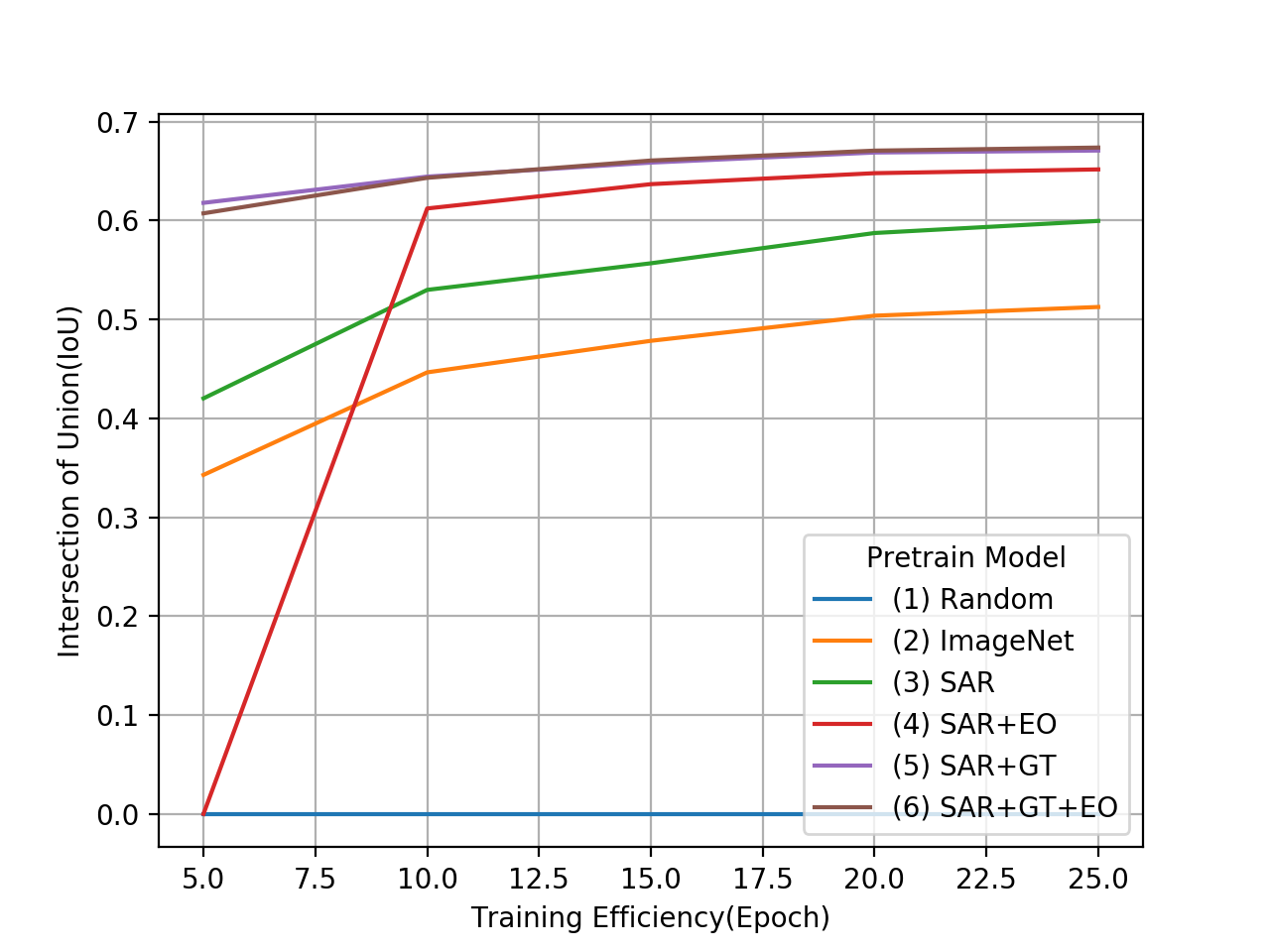} }}%
 \caption{Accuracy (Left) and Building Intersection of Union (Right) according to the number of training epochs with ResNet50 + U-Net segmentation model. Zoom in for best view.}%
 \label{fig:epoch_performance}%
\end{figure}

\begin{table}[]
    \centering
    \begin{adjustbox}{width=0.36\textwidth}
    \begin{tabular}{l|l|c|c}
    \toprule
    Backbone & Pretrain Source & Acc & Building IoU \\
    \midrule
    ResNet50 & (1) Random & $0.9302$ & $0.2732$ \\
         &  (2) ImageNet & $0.9502$ & $0.4896$ \\
         &  (3) SAR & $0.9673$ & $0.6280$ \\
         &  (4) SAR+EO & $0.9736$ & $0.6874$ \\
         &  (5) SAR+GT & $0.9738$ & $0.6878$ \\
         &  (6) SAR+GT+EO & $\bm{0.9738}$ & $\bm{0.6890}$ \\
    \midrule
    ResNext50 & (1) Random & $0.9340$ & $0.3039$ \\
         &  (2) ImageNet & $0.9561$ & $0.5311$ \\
         &  (3) SAR & $0.9649$ & $0.6061$ \\
         &  (4) SAR+EO & $0.9746$ & $0.6945$ \\
         &  (5) SAR+GT & $0.9752$ & $0.7006$ \\
         &  (6) SAR+GT+EO & $\bm{0.9756}$ & $\bm{0.7054}$ \\
    \midrule
    Resnest50 & (1) Random & $0.9411$ & $0.3885$ \\
         &  (2) ImageNet & $0.9599$ & $0.5623$ \\
         &  (3) SAR & $0.9720$ & $0.6687$ \\
         &  (4) SAR+EO & $0.9744$ & $0.6920$ \\
         &  (5) SAR+GT & $0.9749$ & $0.6974$ \\
         &  (6) SAR+GT+EO & $\bm{0.9752}$ & $\bm{0.7017}$ \\
    \midrule
    \end{tabular}
    \end{adjustbox}
    \caption{This table shows the performance of building segmentation according to the conditions of contrastive multiview coding by backbone architecture in the DeepLabV3 segmentation model. The same was applied to ResNet50, ResNext50, and Resnest50 to confirm that contrastive multiview coding shows consistent results for the various backbone.}
    \label{tab:Result Table Backbone}
\end{table}

If a deep neural network achieves a good performance in downstream tasks with fewer training epoch and training samples, the neural network is evaluated as a good representation extractor. Because of the building semantic segmentation dataset used in this study, we compare the semantic segmentation performance in the SAR image domain by the percentage of training data, based on the training epoch according to each pretraining method. When evaluating the performance, if the probability of building is more significant than 0.5, it is defined as a building pixel.

Table \ref{tab:Result Table} shows the performance according to which pretraining method and which segmentation architecture is used. Each segmentation architecture is trained through the same training step as mentioned in Section \ref{sec:exp}. Comparisons within a single segmentation architecture reveal the following. When any pretraining is not performed, the lowest results are shown. Moreover, even if the same number of positive samples are used, it can be seen that the greater the number of types of modality that are used, the better the performance is. In addition, it can be confirmed that these phenomena occur consistently in the three segmentation models.

Table \ref{tab:Result Table Backbone} shows the performance changes according to backbones with the DeepLabV3 segmentation model. Each backbone was pretrained with the same protocol in Section \ref{sec:training-details}. Table \ref{tab:Result Table Backbone} indicates that the effectiveness' tendency of pre-training is similar with the previous discussion regardless of the type of backbone model.

Figure \ref{fig:epoch_performance} shows the semantic segmentation performance according to the training epochs. We can see that the more types of modailities that are used in pretraining, the better the performance that can be obtained with the sample training steps. Specifically, owing to the characteristics of the SAR images, only the cross-sectional information of the building can only be estimated when only the SAR images are used. However, if pretraining is performed using the EO images or a segmentation map, it is possible to estimate the information of various aspects of the building. In addition, it is possible to outperform the ImageNet pretrained model with only a few epochs. Extremely, the model in (6) exceeded the 25 epochs performance of model (2), ImageNet pretrained model, in just five epochs, and in the case of model (4) using only EO without using segmentation map, it only takes 10 epochs to outperform model (2).

\begin{figure}[t]%
 \centering
 \subfloat{{\includegraphics[width=0.25\textwidth]{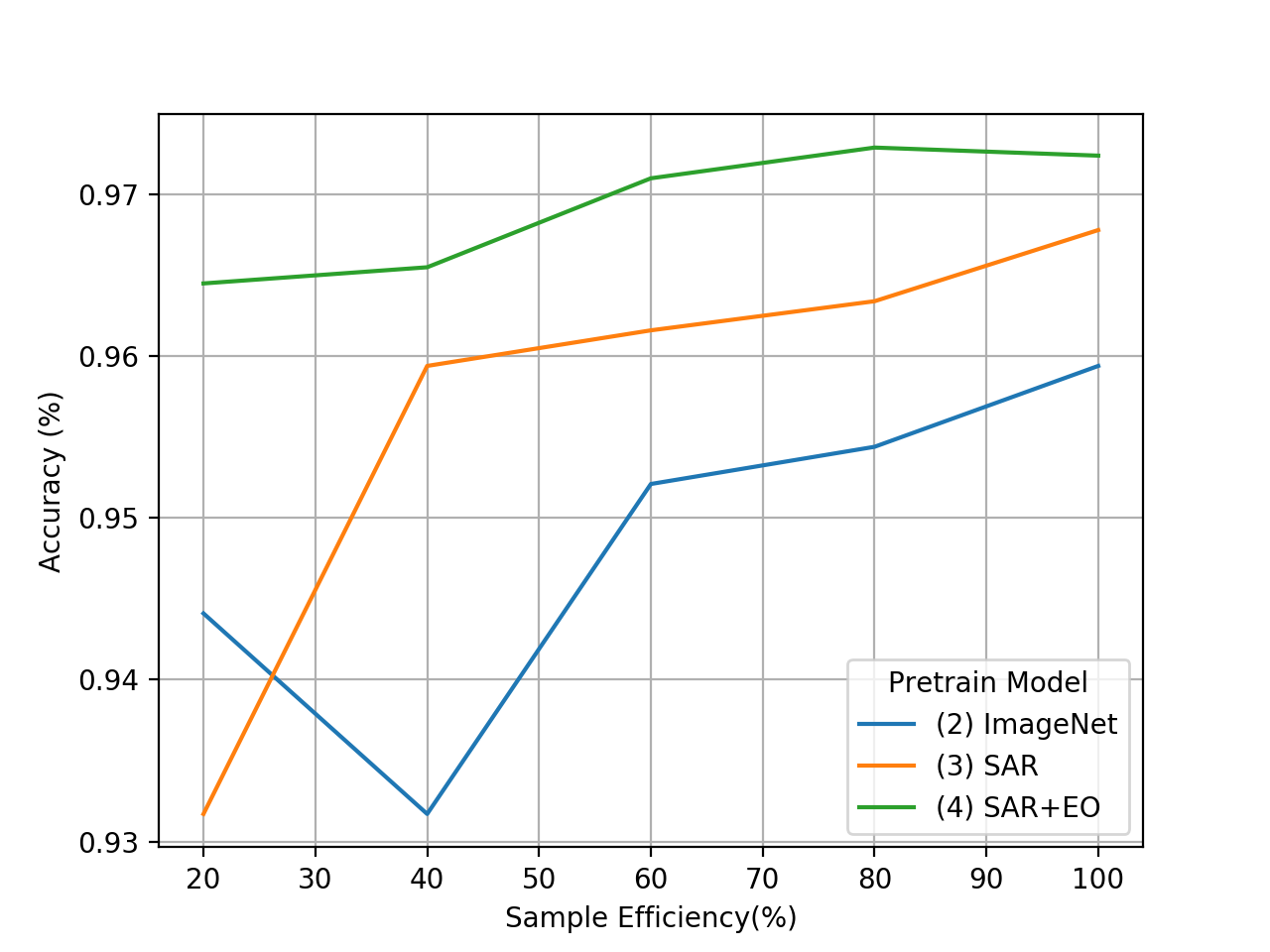} }}%
 \subfloat{{\includegraphics[width=0.25\textwidth]{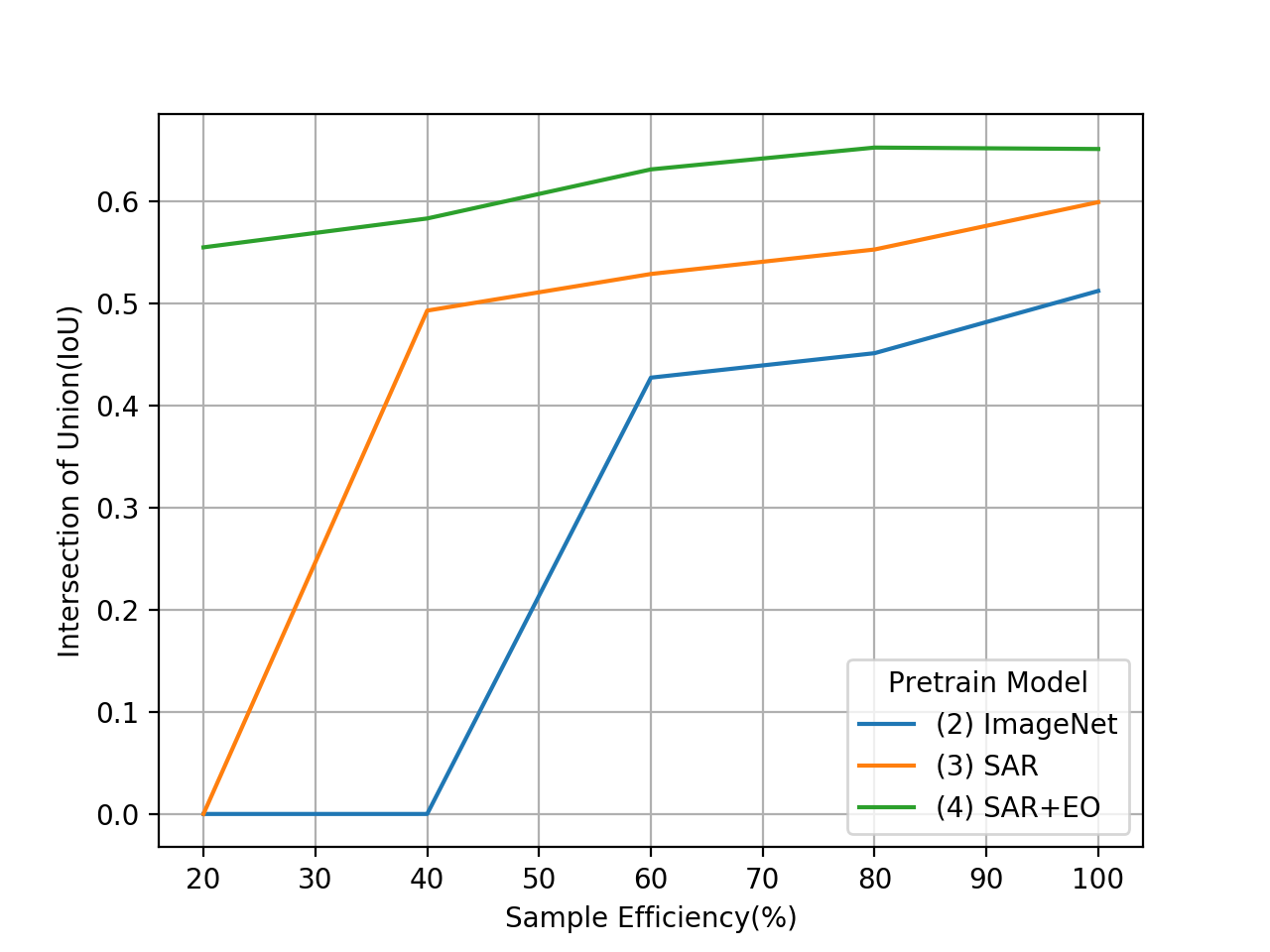} }}%
 \caption{Accuracy (Left) and Building Intersection of Union (Right) according to the proportion of training samples with ResNet50 + U-Net segmentation model. Zoom in for best view.}%
 \label{fig:sample_performance}%
\end{figure}

Figure \ref{fig:sample_performance} shows the performance of each model according to the percentage of training data. Because the performance should be compared according to how much of the segmentation map is used in the training data, both (5) and (6) using segmentation maps during the pretraining process were excluded from the comparison. In addition, the random initialized model (1) is excluded because the performance with any condition is zero. As it can be seen in the figure, even if only 20\% of the training sample is used for learning in (4), a better performance than (2) when using 100\% of the training sample can be obtained. In addition, comparing contrastive coding only with SAR, a better performance is generally obtained when using a large number of modalities.

\section{Conclusion}
\begin{figure*}%
    \centering
    \includegraphics[width=0.9\textwidth]{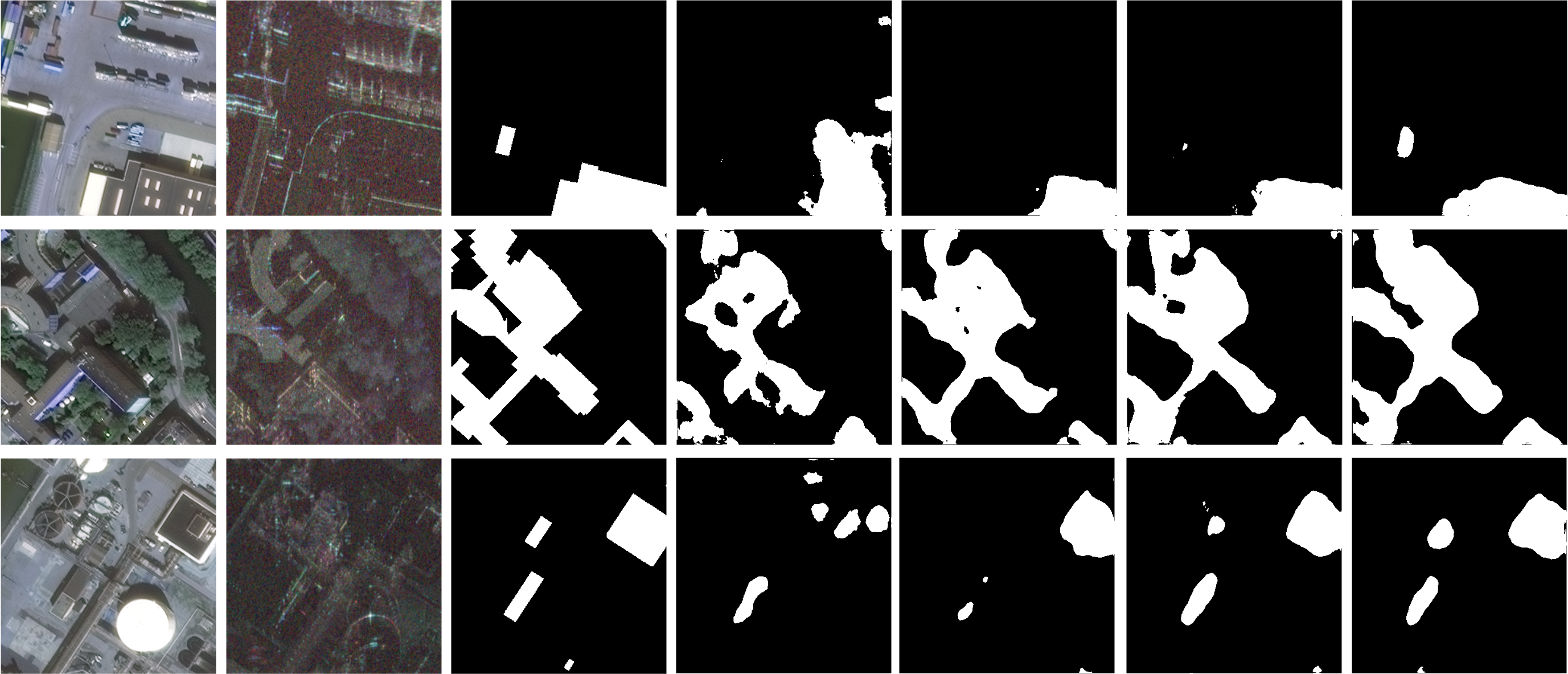}
    \caption{The building semantic segmentation results of ResNet50 + U-Net model from each pretraining scheme. The leftmost three images represent the EO image, the SAR image, and the semantic map in turn. From the next image, four semantic segmentation results according to different model initialization schemes are shown. In order from left, the results of \textit{Random Initialization}, \textit{ImageNet-pretrained Weight}, \textit{CMC with SAR + EO}, \textit{CMC with SAR + GT + EO} are shown.}%
    \label{fig:results}%
\end{figure*}

This letter investigated the multimodal approach to obtain good representation features in the SAR imagery domain.
To evaluate whether a good representation feature is obtained, building semantic segmentation as a downstream task was conducted in the SAR imagery domain, and the performance was compared.
Besides, to compare the learning efficiency, performance comparison according to the number of training epochs and the ratio of the training data was also conducted.

In conclusion, the greater number of modalities that are used, the better representation features that can be obtained when pretraining more positive sample sets, and the better performance can be obtained in downstream tasks. Furthermore, since the characteristics of the SAR imagery domain are extremely different from those of the public image dataset, it can be observed that pretraining improves model performance especially when there are little data available.


%



\ifCLASSOPTIONcaptionsoff
  \newpage
\fi



%
\bibliographystyle{IEEEtran}
\bibliography{bibtex/bib/IEEEexample}

%






\end{document}